# Self-supervised audiovisual representation learning for remote sensing data

Konrad Heidler [a], Lichao Mou [a,*], Di Hu [b,c], Pu Jin [a], Guangyao Li [b,c], Chuang Gan [d], Ji-Rong Wen [b,c], Xiao Xiang Zhu [a,*]

[a] *Technical University of Munich (TUM), Munich, Germany*
[b] *Gaoling School of Artificial Intelligence, Renmin University of China, Beijing, China*
[c] *Beijing Key Laboratory of Big Data Management and Analysis Methods, Beijing, China*
[d] *MIT-IBM Watson AI Lab, Cambridge, MA, 02142, USA*

ABSTRACT

Many deep learning approaches make extensive use of backbone networks pretrained on large datasets like ImageNet, which are then fine-tuned. In remote sensing, the lack of comparable large annotated datasets and the diversity of sensing platforms impedes similar developments. In order to contribute towards the availability of pretrained backbone networks in remote sensing, we devise a self-supervised approach for pretraining deep neural networks. By exploiting the correspondence between co-located imagery and audio recordings, this is done completely label-free, without the need for manual annotation. For this purpose, we introduce the *SoundingEarth* dataset, which consists of co-located aerial imagery and crowd-sourced audio samples all around the world. Using this dataset, we then pretrain ResNet models to map samples from both modalities into a common embedding space, encouraging the models to understand key properties of a scene that influence both visual and auditory appearance. To validate the usefulness of the proposed approach, we evaluate the transfer learning performance of pretrained weights obtained against weights obtained through other means. By fine-tuning the models on a number of commonly used remote sensing datasets, we show that our approach outperforms existing pretraining strategies for remote sensing imagery. The dataset, code and pretrained model weights are available at https://github.com/khdlr/SoundingEarth.

## 1. Introduction

Imagine yourself standing in a lush green forest. You can see the green of the trees around you, maybe a brown, muddy path below your feet. At the same time, you can hear the leaves rustling in the wind, and the songs of some birds nearby. Now try to imagine one without the other, the same forest scenery but completely silent, or the soundscape without any visual context. Chances are, you will find it hard to clearly separate these impressions completely.

In most situations, our mind makes use of multiple of our senses to perceive the scenery around us. By basing our perception of the world on multiple senses, we get a more robust impression of our surroundings than if we were to rely on a single sense. In fact, phenomena like the *McGurk effect* (McGurk and MacDonald, 1976) suggest that the distinction between human vision and hearing might not even be as clear as we think .

Given the great added value of combining our vision and hearing as stated above, the simultaneous processing of visual imagery and sounds is something that comes very natural to us as humans. Recent studies have shown remarkable advances in audiovisual machine learning (Arandjelovic and Zisserman, 2017; Petridis et al., 2018; Tzirakis et al., 2017). However, there still remains a paucity of research on understanding the earth through multiple modalities at once (Hu et al., 2020a). In order to enable such research, we present the large-scale *SoundingEarth* dataset that consists of crowdsourced audio and aerial imagery captured at the same geographical location, and is tailored towards audiovisual learning in the context of remote sensing (see Fig. 1).

At the same time, the pre-training of deep learning models in remote sensing is a topic of high importance that needs specialized approaches due to the large number of different imaging platforms and tasks in the field. We show that the task of matching imagery and audio is instructive for neural networks in the sense that it teaches the networks how to learn useful and general features without the need for labels. This is true even when disregarding the auditive information in later downstream applications, which implies that the audiovisual correspondence also helps the model better understand the image modality by itself.

* Corresponding authors.
*E-mail addresses:* lichao.mou@tum.de (L. Mou), xiaoxiang.zhu@tum.de (X.X. Zhu).

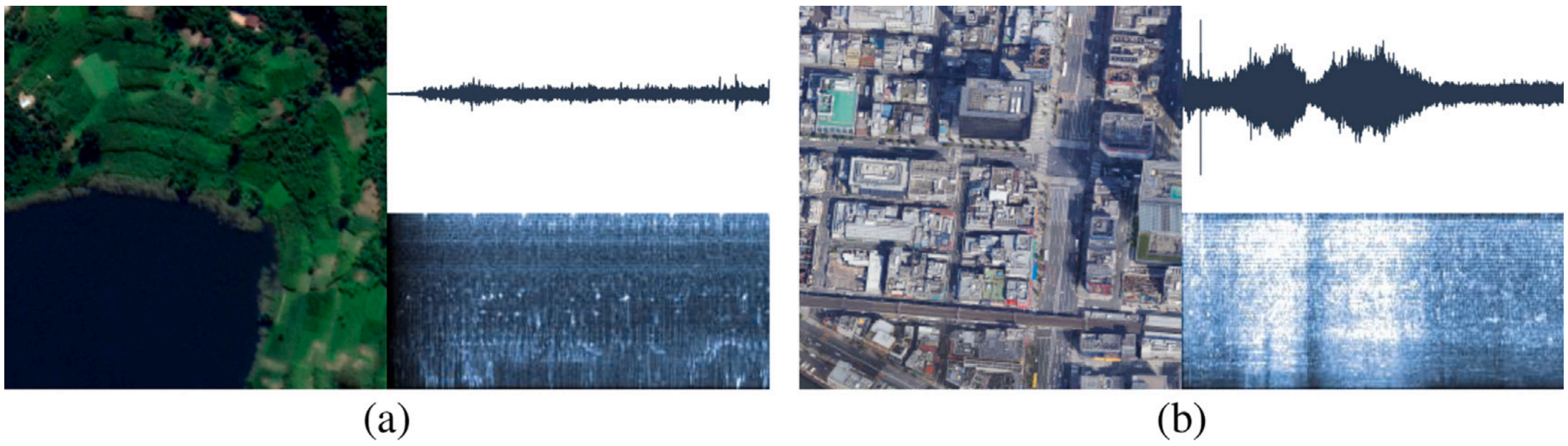

**Fig. 1.** Two examples of corresponding imagery and audio: Lake Bunyonyi, Uganda (a) and a city scene in Tokyo, Japan (b). Visualized for both scenes is the aerial image (left), the audio waveform (top right) and the extracted log-mel spectrogram (bottom right). For visualization purposes, the audio data was clipped to the first 20 s, even though the full samples are much longer. The audio can be listened to at `archive.org/details/aporee_41512_47342`, and `archive.org/details/aporee_46891_53254`, respectively.

To our knowledge, this is the first work to introduce self-supervised pretraining from scratch on audiovisual remote sensing data. As we show in our experiments, network weights trained in this way are better suited for a number of downstream tasks than those obtained by self-supervised pretraining on the single modality of aerial imagery or the commonly used ImageNet weights.

To summarize, this work's contributions are threefold.

- In Section 3, we build a large-scale dataset to facilitate this task, called *SoundingEarth*, which consists of more than 50k co-localized field recordings and overhead imagery pairs, collected from a publicly available audio source.
- In Section 4, we describe our framework for the pretraining of deep neural network models based on the audiovisual correspondence of aerial imagery and field recordings. This framework is independent of the type of imagery used and can therefore be applied to arbitrary types of remote sensing imagery. It is trained using a batch triplet loss, which we propose to combine the benefits of classical triplet loss training with those of recent contrastive learning methods.
- In Section 5, we report and discuss the results of our extensive experiments on downstream tasks that demonstrate the effectiveness of our approach with superior performance over state-of-the-art methods.

## 2. Related work

### 2.1. Audiovisual learning

Exploiting the relationship between audio and imagery is an emerging topic in the machine learning community. Pioneering works addressed tasks like audiovisual speech recognition (Petridis et al., 2018) and affect classification (Soleymani et al., 2011; Tzirakis et al., 2017). In the deep learning era, their relationship is further investigated in cross-modal transfer learning, where the predictions of a well-trained network of one modality are employed as the supervision for training a student network for the other modality (Aytar et al., 2016; Owens et al., 2016b).

Quite a few approaches directly use massive unlabeled video datasets for self-supervised model training. Arandjelovic and Zisserman (2017) train a model on the frame-to-sound correspondence. Without any additional supervision, this model gains the ability to discern semantic concepts in both modalities. Similarly, deep clustering approaches can also learn meaningful representations when using the clustering information as supervision for the other modality (Alwassel et al., 2020). Other tasks include temporal alignment (Korbar et al., 2018; Owens and Efros, 2018), sound source localization (Senocak et al., 2018; Qian et al., 2020) and sound separation (Zhao et al., 2018; Gao and Grauman, 2019).

Collecting video data tailored towards a specific task is also an option. Owens et al. (2016a) recorded short video clips where they hit or scratched a large variety of objects with a drumstick. They then trained a network to predict the resulting sounds from just the visual video data, as well as to predict the material of the probed object from both video and audio. Apart from directly using both modalities, Using derived data as a target, Owens et al. (2016b) predict audio statistics for the ambient sound from an image. Going even further, text data can be added as a third modality (Aytar et al., 2017).

Other than these works, this paper explores the audiovisual relationship in terms of geographical location, based on which we aim to understand the earth in an audiovisual way.

A few existing works address audiovisual machine learning in the context of remote sensing. Salem et al. (2018) propose to combine the audiovisual correspondence with a clustering algorithm to build an "aural atlas". Further, Hu et al. (2020b) shows that fusing audiovisual information can greatly benefit the task of crowd counting. Finally, retrieving data in a cross-modal setting is an emerging task in remote sensing that aims to help users experience the world in an audiovisual way (Mao et al., 2018; Chen and Lu, 2020; Chen et al., 2020b).

### 2.2. Pre-trained models in remote sensing

Aside from self-supervised learning, there are a number of methods for initializing deep learning models for remote sensing tasks before training, which we will explain here in brief.

*Random initialization.* Perhaps the most common approach is to initialize models from scratch, i. e. initializing the weights in a completely random fashion and only training the model on the given dataset. While random model initialization can be used for data from all sensing platforms, it requires a large amount of labeled data to converge to satisfactory results, and can lead to overfitting and poor generalization (Zhu et al., 2017).

*ImageNet weights.* When working with RGB data, models pretrained on natural imagery tasks, like the ImageNet dataset (Deng et al., 2009) can be used. Even though the modalities of ground-level and overhead imagery are very different from each other, this approach sometimes works surprisingly well (Guo et al., 2020). However, most remote sensing platforms have more spectral channels than RGB, or even use other acquisition modes than optical imaging, like synthetic aperture radar.

*Specific remote sensing datasets.* For some satellite sensors, large labeled datasets are available for pretraining. For the Sentinel-1 and -2 satellites, BigEarthNet (Sumbul et al., 2021) matches a large imagery archive with corresponding land cover annotations. Other comparable datasets include DENETHOR for crop type mapping (Kondmann et al., 2021), and Million-AID for aerial scene classification (Long et al., 2021). These datasets are usually specific to a certain sensor and application, e. g. land cover mapping with Sentinel-2 imagery. Approaches to making such models more general or transferring them to different applications and sensors exist. For an in-depth study of current methods for this, the reader is referred to the recent review study by Qin and Liu (2022).

**Table 1**
Comparison of audiovisual datasets focusing on remote sensing imagery.

| Dataset | Samples | Audio | | Imagery | |
|---|---|---|---|---|---|
| | | Source | Duration | Source | Resolution |
| CVS (Salem et al., 2018) | 23 308 | Freesound | N/A | Bing Maps | 0.6 m |
| ADVANCE (Hu et al., 2020a) | 5 075 | Freesound | ~14 h | Google Earth | 0.4 m |
| SoundingEarth | 50 545 | radio aporee | ~3500 h | Google Earth | 0.2 m |

### 2.3. Self-supervised model pre-training

Recently, self-supervised learning methods like MoCo (He et al., 2020), SimCLR (Chen et al., 2020a), BYOL (Grill et al., 2020) and SwAV (Caron et al., 2020) have significantly reduced the gap with supervised methods on natural images. The most recent self-supervised models pretrained on ImageNet even surpass supervised pretrained models on multiple downstream tasks (He et al., 2020).

By their very nature, self-supervised methods can usually be applied to any sensor and derive general features not tailored for a specific application, making them strong candidates for the pretraining of deep neural networks for remote sensing tasks. Stojnić and Risojević (2018a,b) used split-brain autoencoders to analyze aerial images, and explored the number of images used for self-supervised learning and the influence of the use of different color channels on aerial image classification. Ayush et al. (2020) introduced a contrastive loss and a loss term based on image geolocation classification. Tao et al. (2020) analyzed the possibility of using different self-supervised methods, like image restoration, context prediction and conducted training on a small remote sensing image dataset of 30,000 images. Additionally, Kang et al. (2020) trained on 100,000 remote sensing image patches using comparative learning with different enhancements and tested them on the NAIP (Jean et al., 2019) and EuroSAT (Helber et al., 2019) tasks.

These self-supervised learning methods have made a series of achievements in remote sensing data analysis, but they only consider information from the visual mode, and do not use information from other modalities. In this work, we utilize self-supervised audiovisual representation learning for downstream tasks on aerial imagery.

Early works include greedy layer-wise unsupervised pretraining (Romero et al., 2016). and Tile2Vec (Jean et al., 2019), where the model was trained to match imagery patches based on spatial proximity, inspired similar pretraining tasks for natural language data. Recent studies in pretraining for remote sensing often involve some pre-text tasks like colorization of images (Vincenzi et al., 2020), super-resolution (Peng et al., 2021), or overlap detection (Leenstra et al., 2021). It is also possible to pre-pretrain a network on natural imagery before pretraining on aerial imagery in a second step (Reed et al., 2021).

Exploiting specific properties of remote sensing data for self-supervised learning is another promising area of research that is quickly gaining traction. Ayush et al. (2020) suggested a way of extending MoCo (He et al., 2020) with a geography-aware loss term, which improves the learned representations compared to training using regular MoCo. Going in a different direction, Mañas et al. (2021) combine acquisitions from different seasons with traditional image augmentations, and encode the information in multiple orthogonal subspaces. Similar to these works, our goal is to leverage the colocation of audio and imagery to learn features valuable features.

## 3. The *SoundingEarth* Dataset

We introduce a dataset for geo-aware audiovisual learning in remote sensing, which we call the *SoundingEarth* Dataset. The development of the dataset is split in two steps, the acquisition and cataloging of geo-tagged audio data, and the extraction of corresponding overhead imagery.

### 3.1. Collection of geo-tagged audio

Sources for representative and geo-tagged audio are rare. Among the few public audio libraries that include geo-tags, most contain samples that have little connection to their geographical surroundings. In contrast, gathering audio samples that capture a local ambience is the central point of the radio aporee ::: maps project (Noll, 2019). Started in 2006 by Udo Noll, the project represents a crowd-sourced effort of collecting a global soundmap from many geo-tagged *field recordings*, which refer to any audio recordings made “in the field”.

Anyone can contribute to this soundmap by uploading their own recordings, yielding a nearly global coverage of the samples, as can be seen in Fig. 2. The guidelines for uploading sounds to the site include requirements for quality, length, and a focus on local ambience. Upon uploading, the creators put their recordings under one of the creative commons licenses, or release them into the public domain, making the audio data fit as a training set for machine learning approaches. All of radio aporee ::: maps’s audio data is mirrored on the *Internet Archive*.[1] Owing to the clear orientation of radio aporee ::: maps towards field recordings most of the recorded audio samples give the listener a vivid impression of the recorded scene. For geospatial analysis, this project therefore constitutes a treasure trove of audio data.

At the time of our download, the database contained about 435 GB of high quality audio data, with metadata for each sample including the geographical coordinates, the creator’s name necessary for correct attribution, and in many cases a short textual description of the audio.

### 3.2. Collection of aerial imagery

Using the geographical coordinates from the audio samples, we matched the audio samples with corresponding imagery by extracting image tiles from Google Earth in an automated fashion. Given the longitude and latitude where the audio was recorded, a tile of $1024 \times 1024$ pixels is extracted at the highest available resolution from Google Earth. This implies a spatial resolution of approximately 0.2 m per pixel.

### 3.3. Data cleaning

As already mentioned, the audio recordings in the dataset have an exceptional level of quality, both regarding audio fidelity and the recorded content. Therefore, few manual corrections were needed. As it is infeasible to listen to the thousands of hours of audio content, our data cleaning routine was limited to a full-text search over the recordings’ filenames and textual descriptions to filter out nondescript audio samples like “testsound.mp3”. During this semi-automated cleaning process, 621 samples were excluded from the dataset.

[1] The collection can be found under the following link: https://archive.org/details/radio-aporee-maps.

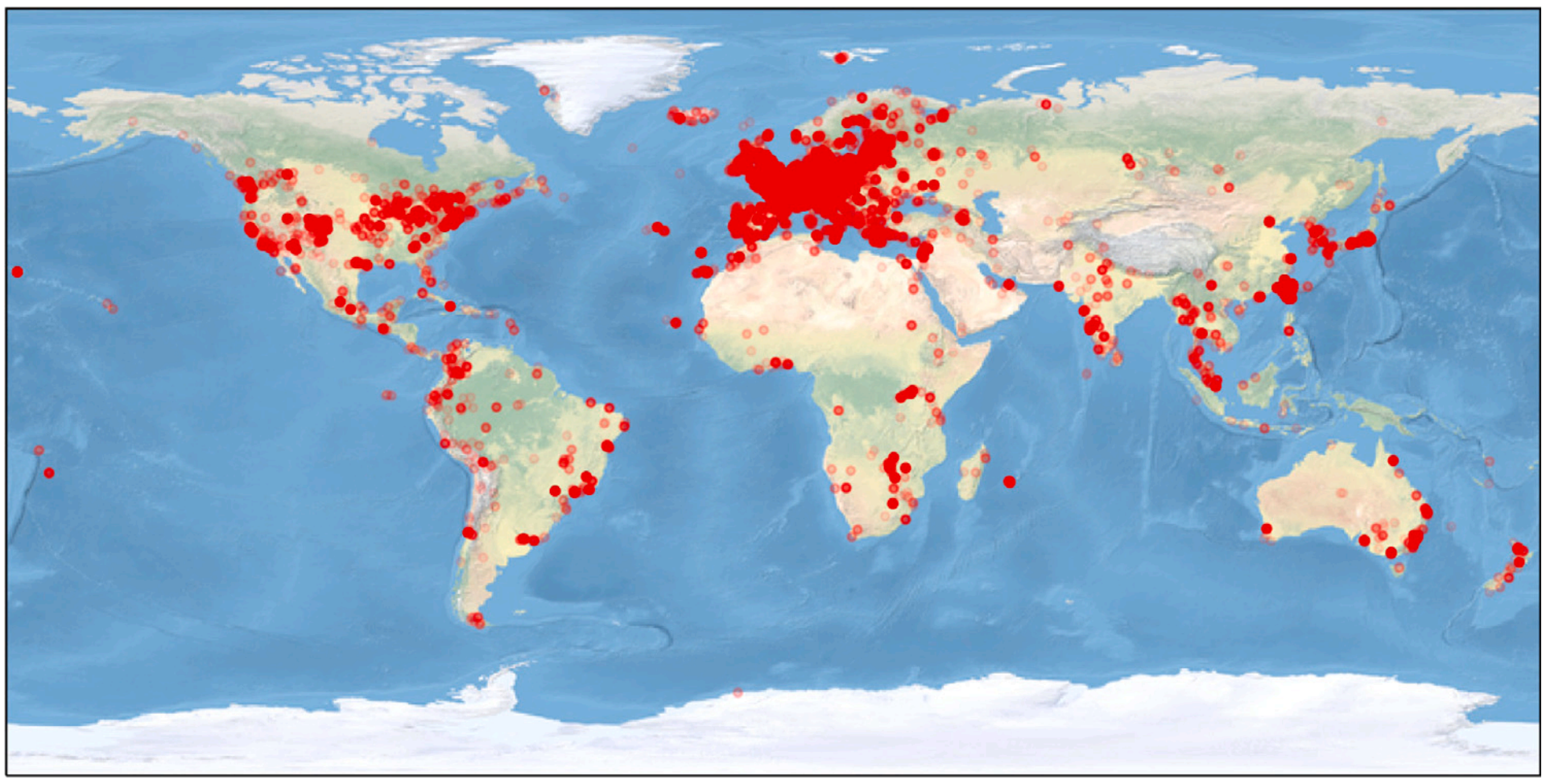

**Fig. 2.** Spatial distribution of samples in our SoundingEarth dataset.

### 3.4. Dataset overview and discussion

At the time of our download radio aporee ::: maps had collected over 50,000 geo-tagged field recordings from 136 countries all over the world, as shown in Fig. 2. As a result, our built *SoundingEarth* dataset consists of 50,545 image-audio pairs. The total length of the audio amounts to more than 3500 h of ambient sounds, making the dataset much larger than existing audiovisual datasets focusing on aerial imagery (see Table 1).

One notable property of the dataset is the extreme skew of audio durations. While the median duration is about 3 min, the longest 1% of the audio samples exceed half an hour in duration. The general distribution of the duration in minutes is shown in Fig. 3.

The aerial images were all collected during the daytime. Meanwhile, some of the audio recordings (~13%) were also collected at night, which might lead to a distributional shift between the two modalities. For self-supervised learning however, this does not appear to pose a problem. On the contrary, we believe that the nighttime recordings can act like another type of auditive view of the given location, much like data augmentations act for other self-supervised learning frameworks. As the recording time is available in the metadata, it is easy to filter out nighttime recordings if required.

As with many crowd-sourced datasets, the geographic distribution of the samples is biased towards regions with many users (cf. Fig. 2). While the dataset has samples from many places in the world, there is a strong bias towards some regions. For instance, 86% of all samples were recorded in either Europe or Asia, while less than 2% of the samples are from Africa.

To facilitate further research in audiovisual based geo-understanding, the *SoundingEarth* dataset is publicly available (Heidler et al., 2021).

## 4. Audiovisual model pre-training

Following recent advances in self-supervised learning for images (Chen et al., 2020a; He et al., 2020), we develop a framework to automatically learn representations from paired audiovisual data. The goal of this framework is to build a common embedding space for imagery and audio, where the embeddings of corresponding audiovisual pairs are close together while the embeddings for distinct pairs are farther apart from each other. For both modalities, we train a CNN to perform this projection. The underlying assumption of this methodology is the idea that the networks will learn features that represent the commonalities between the visual imagery and the sound recorded at the scene. In turn, these features need to be of a high abstraction level, and will therefore be useful for a number of downstream tasks (see Fig. 4).

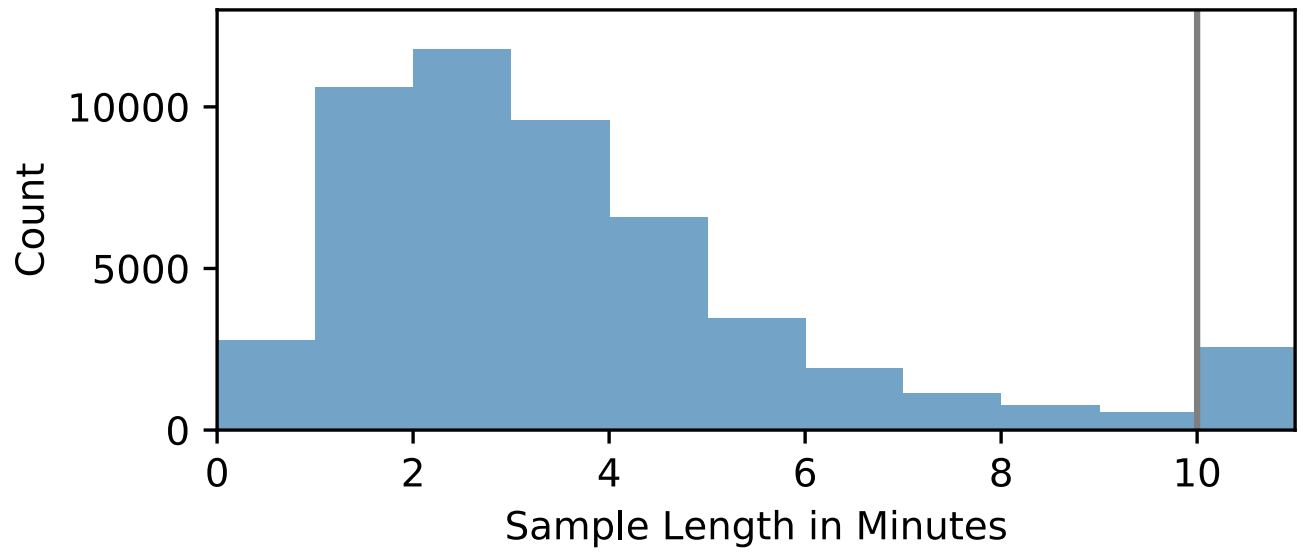

**Fig. 3.** Histogram of the audio durations in the dataset. The rightmost bin sums up all durations longer than 10 min.

### 4.1. Data preparation and augmentation

Before training the networks, the input data is transformed into a suitable format for the CNNs. While imagery is the natural input domain for CNNs, digital audio is represented as a waveform, which consists of a sequence of samples. To get the audio into a more serviceable representation, we first apply a short-time Fourier transform (STFT), which converts the audio into a two-dimensional representation. The added second dimension represents the audio frequency. The squared absolute values of these coefficients are then mapped to mel-scale using 128 filter bands. Finally, the logarithm is taken to arrive at a log-mel spectrogram. After this conversion process, the audio representation is equivalent to that of a grayscale image with size $128 \times T$, where $T$ depends on the duration of the audio sample.

In order to prevent pure memorization of the input data and introduce more variety into the training samples, a number of data augmentation techniques are applied both to the imagery as well as the audio spectrograms.

Given the arbitrary length of the audio, a random sample of 128 consecutive spectrogram frames is extracted from the original spectrogram, resulting in a square spectrogram of size $128 \times 128$. Further audio augmentations like random volume adjustments or frequency shifts did not improve training results, likely because of the translation-invariance of CNNs and the scale-invariance introduced by batch normalization layers.

For the training images, we first cropped the central half of the image to ensure that the augmented scenes do not deviate too far from the true location. Then, a square crop sized randomly between 192 and 384 pixels was extracted and scaled to 192 pixels in size. This sampling strategy ensures that the true source of the audio is within the cropped region around 90% of the time. As the source of the audio can be at any position in the cropped image, the model is forced to extract image features of high generality, capturing all possible audio sources within the image.

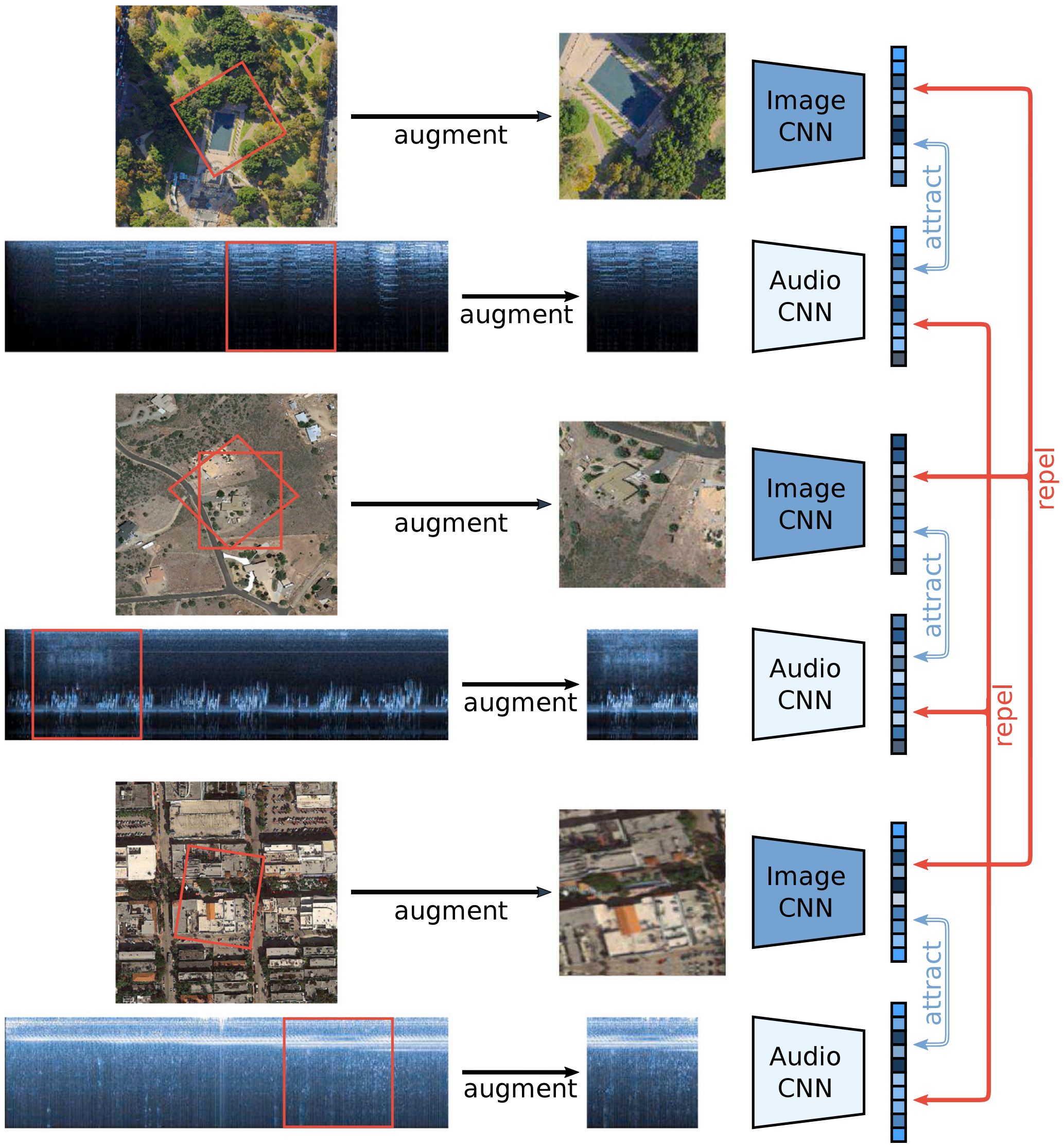

**Fig. 4.** Overview of the proposed pretraining method. After sampling a batch, the corresponding images and spectrograms are augmented = and then embedded into the representation space by the Image and Audio CNNs. The loss function then causes corresponding image and audio embeddings to be drawn together, while samples from different locations are pushed away from each other.

Finally, random adjustments were done with regard to rotation, blur, hue, saturation and value (lightness). Efficient data augmentations were enabled by the *albumentations* python library (Buslaev et al., 2020).

### 4.2. Embedding networks

Pixel images and sound waveforms exist in distinct representation spaces and have different statistical properties (Srivastava and Salakhutdinov, 2014). In order to still derive common features and represent these highly non-linear semantic correlations across modalities, image and sound are encoded with modality-specific networks to represent them in the common embedding space.

#### 4.2.1. Visual subnet

The visual pathway adopts a ResNet architecture (He et al., 2016). It is working on inputs of size $3 \times 192 \times 192$ at training time. To better assess the transferability of the framework to different network architectures, both ResNet-18 and ResNet-50 are evaluated as visual encoders. After the convolutional stages of the ResNet, the data is transformed into rich feature maps. In order to get a single vector representing the entire image, these feature maps are merged using global average pooling, followed by a final fully connected layer.

#### 4.2.2. Audio subnet

The audio pathway operates on log-mel spectrograms of size $1 \times 128 \times 128$. Given the reduced complexity of the spectrograms compared to RGB imagery, we only employ a ResNet-18 encoder for this subnet. Just like with the visual encoder, the convolutional features of the ResNet encoder are globally averaged and fed into a final fully connected layer, so that the final representation for both modalities is given by a single vector for each.

### 4.3. Batch triplet loss

After acquiring representations for visual and audio inputs, we train the two networks in a way that encourages corresponding bimodal inputs to match each other closely in the embedding space.

Conventional representation learning methods compare the embeddings of two (Hadsell et al., 2006) or three (Weinberger and Saul,

**Algorithm 1** Batch-wise Triplet Loss in PyTorch-like pseudocode.

```
def batch_triplet_loss(v, a):
  diff = v.unsqueeze(1) - a.unsqueeze(0) # pairwise
      differences
  D = norm(diff, dim=2) # distance matrix
  D_true = D.diagonal() # distances of the true pairings

  d_col = sum(relu(D_true.unsqueeze(0) - D + 1.0))
  d_row = sum(relu(D_true.unsqueeze(1) - D + 1.0))

  return d_col + d_row
```

2009; Schroff et al., 2015) samples, which discards a lot of possible learning feedback. Therefore, a key idea in recent contrastive learning techniques is to use all possible pairings in a training batch (Chen et al., 2020a).

We combine this idea with triplet loss, resulting in a *batch triplet loss* objective. For visual embeddings $v_i$ and corresponding audio embeddings $a_i$, we first calculate the matrix of pairwise distances $D(a, v)$ as

$$D = \begin{pmatrix} \|a_1 - v_1\|_2 & \dots & \|a_1 - v_n\|_2 \\ \vdots & \ddots & \vdots \\ \|a_n - v_1\|_2 & \dots & \|a_n - v_n\|_2 \end{pmatrix}. \quad (1)$$

The objective of the representation learning procedure should then be to minimize the diagonal entries of that matrix while keeping all other values above a certain margin. Keeping in mind the original formulation of the triplet margin loss function (Weinberger and Saul, 2009) as

$$\mathcal{L}(x, y^+, y^-) = \max(0, \|x - y^+\|_2 - \|x - y^-\|_2 + 1), \quad (2)$$

we apply this to all possible pairings of diagonal elements and off-diagonal elements for each row and column:

$$\begin{aligned} \mathcal{L}(D) = &\sum_i \sum_{j\neq i} \max(0, D_{ii} - D_{ij} + 1) \\ &+ \sum_j \sum_{i\neq j} \max(0, D_{ii} - D_{ij} + 1). \end{aligned} \quad (3)$$

Algorithm 1 shows the pseudocode for this loss function. Our experiments in Section 5.6 show that this approach outperforms the naive triplet loss formulation from Eq. (2) as well as the contrastive loss used in Chen et al. (2020a) for most downstream applications.

### 4.4. Applicability to other sensors

The training procedure does not make use of any specific properties of the RGB imagery used, and is therefore largely agnostic of the imagery used. It should be very straightforward to extend to other earth observation sensors like Sentinel-1, Sentinel-2 or Landsat by changing the number of input channels to the visual sub-network. As these platforms take images at a much lower resolution than the aerial images used for this study, we recommend reducing the extracted patch size from 1024 × 1024 to a lower value like 128 × 128.

The comparison with existing methods for these sensors (cf. Section 2.2) should open up exciting avenues for future work.

## 5. Transfer learning experiments

The penultimate goal of this work is to provide pretrained networks for downstream applications. In order to confirm the hypothesis that our weights are indeed better suited for remote sensing tasks than other sets of weights, we evaluate them against a number of competitors on several downstream tasks.

### 5.1. Competing pretraining schemes

#### 5.1.1. Random

The predominant method of initializing backbone weights in remote sensing is to initialize them completely at random. To quantify the benefit of pretrained weights, we evaluate random weight initialization as a baseline.

#### 5.1.2. ImageNet

The first actual pretraining method for RGB imagery is to use weights trained on the classification task in the ImageNet dataset (Deng et al., 2009). As these are readily available in most deep learning frameworks, this method is very common and has proven successful on ground-level imagery and some remote sensing tasks as well. However, due to the different nature of ImageNet images and remote-sensing overhead images, we speculate that this might not be the optimal strategy.

#### 5.1.3. Tile2Vec

This method learns weights in a self-supervised fashion from the spatial relations of overhead imagery (Jean et al., 2019). In the original paper, the weights were trained on NAIP imagery which includes not only RGB but an additional NIR channel as well. For our experiments on RGB imagery, we have to impute this fourth channel by the mean value of the other channels, which may lead to decreased performance. As the authors provide only pretrained ResNet-18 weights, this approach is not included in the ResNet-50 evaluations.

#### 5.1.4. Contrastive

Plain contrastive learning without an additional projecting head, as outlined in Chen et al. (2020a).

#### 5.1.5. SimCLR

A recent advance in self-supervised learning was given by SimCLR (Chen et al., 2020a), which combines extensive data augmentation strategies with the contrastive loss objective function.

#### 5.1.6. Momentum Contrast (MoCo)

Another fairly recent addition to the family of self-supervised learning methods, MoCo (He et al., 2020) aims to align the representations of different image augmentations between the model and a *momentum encoder*, which is a copy of the model that is updated via exponential moving average.

For ImageNet weights, we used the ones distributed by the torchvision python package, while for Tile2Vec we used the weights made available by the authors (Jean et al., 2019). All other methods were trained by us on the previously introduced dataset. Naturally the image-only pretraining methods were applied to only the visual part of the data.

### 5.2. Aerial image classification

A common task in remote sensing is to categorize scenes into one of several pre-defined classes. Due to the importance of this task, a great number of available datasets exist. For our comparisons, we have evaluated the models on three such datasets. As the evaluated networks are already pretrained, we follow the evaluation protocol proposed by Xia et al. (2017), where only 50% of the data are used for training and the other 50% are used for evaluation purposes.

**Table 2**
Results on UC Merced (Yang and Newsam, 2010), values displayed in %.

| Weights | Backbone | Accuracy after | | |
|---|---|---|---|---|
| | | 1 epoch | 2 epochs | 5 epochs |
| Random | ResNet-18 | 12.10 | 42.57 | 45.81 |
| ImageNet | ResNet-18 | 46.29 | 59.24 | 82.10 |
| Tile2Vec (Jean et al., 2019) | ResNet-18 | 38.67 | 59.05 | 74.38 |
| Contrastive (Chen et al., 2020a) | ResNet-18 | 39.90 | 63.43 | 80.95 |
| SimCLR (Chen et al., 2020a) | ResNet-18 | 58.95 | 77.33 | 88.48 |
| MoCo (He et al., 2020) | ResNet-18 | 50.86 | 67.05 | 77.33 |
| Ours | ResNet-18 | **71.33** | **85.81** | **90.19** |
| Random | ResNet-50 | 9.24 | 19.71 | 44.95 |
| ImageNet | ResNet-50 | 24.29 | 37.52 | 80.19 |
| Contrastive (Chen et al., 2020a) | ResNet-50 | 39.81 | 67.52 | 84.57 |
| SimCLR (Chen et al., 2020a) | ResNet-50 | 56.48 | 75.71 | 85.43 |
| MoCo (He et al., 2020) | ResNet-50 | 53.71 | 64.29 | 78.95 |
| Ours | ResNet-50 | **72.29** | **87.24** | **89.71** |

**Table 3**
Results on NWPU-RESISC45 (Cheng et al., 2017), values displayed in %.

| Weights | Backbone | Accuracy after | | |
|---|---|---|---|---|
| | | 1 epoch | 2 epochs | 5 epochs |
| Random | ResNet-18 | 31.21 | 42.32 | 57.65 |
| ImageNet | ResNet-18 | 69.83 | **77.89** | **83.76** |
| Tile2Vec (Jean et al., 2019) | ResNet-18 | 52.44 | 58.03 | 69.73 |
| Contrastive (Chen et al., 2020a) | ResNet-18 | 59.41 | 67.75 | 81.49 |
| SimCLR (Chen et al., 2020a) | ResNet-18 | 69.77 | 73.68 | 80.36 |
| MoCo (He et al., 2020) | ResNet-18 | 51.94 | 64.09 | 78.28 |
| Ours | ResNet-18 | **73.82** | 76.30 | 81.71 |
| Random | ResNet-50 | 25.96 | 36.89 | 42.48 |
| ImageNet | ResNet-50 | 68.49 | 72.36 | 83.06 |
| Contrastive (Chen et al., 2020a) | ResNet-50 | 63.55 | 70.60 | 81.34 |
| SimCLR (Chen et al., 2020a) | ResNet-50 | 68.14 | 75.16 | 80.69 |
| MoCo (He et al., 2020) | ResNet-50 | 56.39 | 64.70 | 76.98 |
| Ours | ResNet-50 | **77.17** | **79.82** | **84.88** |

#### 5.2.1. UC merced land use

The first dataset (Yang and Newsam, 2010) contains 2100 overhead images from 21 land-use classes. Images in this dataset are 256 × 256 pixels in size with a spatial resolution of ~0.3 m. Extracted from the USGS National Map Urban Area Imagery collection, they cover various regions in the United States. Results for this dataset are presented in Table 2. Here, our pretraining method clearly shows superior results compared to the other evaluated methods. However, this dataset is sometimes criticized for being both very small and simple to solve (Cheng et al., 2017; Xia et al., 2017). Therefore, we conduct further evaluations on two other datasets which both set out to address these two issues.

#### 5.2.2. NWPU-RESISC45

Created in an attempt to improve upon the size and diversity of the UC Merced dataset, the dataset by the Northwestern Polytechnical University (Cheng et al., 2017) consists of 31,500 images from 45 categories. These images are taken from Google Earth and also have a size of 256 × 256. Other than with the UC Merced dataset, these scenes are of varying resolution (between 0.2 and 30 m per pixel) and are taken from locations all around the world. As can be seen in Table 3, this benchmark task does indeed pose a bigger challenge to the models than the previous one. Both the ImageNet and SimCLR weights are strong contenders on this dataset. However, our method performs on par with these approaches, and even has a slight advantage on the ResNet-50 evaluation.

#### 5.2.3. AID

Much like the NWPU-RESISC45 dataset, the Aerial Image Dataset (AID) by Xia et al. (2017) aims to provide an aerial scene classification dataset that is both large and diverse. It is composed of 10 000 images from 30 categories, which were acquired from Google Earth at varying resolution levels between 0.5 and 8 m per pixel, making it comparable to NWPU-RESISC45 in terms of data modality and size. The main difference here is the fact that the images in AID are 600 × 600 pixels in size, allowing for a larger spatial context window for the scenes.

Also for this benchmark, our method outperforms the competing methods (see Table 4). The ImageNet weights are very far behind on this evaluation, which is surprising given their very good performance on the previous NWPU-RESISC45 dataset. We speculate that the larger image size in this dataset favors those methods actually pretrained on remote sensing imagery, whereas ImageNet consists of ground-level imagery.

### 5.3. Aerial image segmentation

Recent image segmentation approaches make ample use of the deep features provided by backbone networks. To demonstrate that the learned weights in our models do not only capture information from the entire scene, but also local information needed for accurate segmentation, we evaluate a semantic segmentation benchmark as well. The DeepGlobe Land Cover Classification Challenge (Demir et al., 2018) aims to provide a benchmark for this task. It consists of 1146 satellite images that are 2448 × 2448 pixels in size at a pixel resolution of 0.5 m per pixel, covering an area of around 1700 $km^2$. Again, we conduct a fine-tuning benchmark on this dataset where the pretrained models are used as backbones for a DeepLabv3+ model (Chen et al., 2018) for 5 epochs.

For this segmentation task, the modern self-supervised methods are all very close to each other in terms of performance (see Table 5), outperforming the random and ImageNet baselines. In the ResNet-18 benchmark, our audiovisually trained models lead the benchmark. When evaluating ResNet-50 models, both our models and the contrastively pretrained models are both strong contenders with the former showing stronger overall accuracy and the latter scoring a better mIoU.

**Table 4**
Results on AID (Xia et al., 2017), values displayed in %.

| Weights | Backbone | Accuracy after | | |
|---|---|---|---|---|
| | | 1 epoch | 2 epochs | 5 epochs |
| Random | ResNet-18 | 16.32 | 34.04 | 47.24 |
| ImageNet | ResNet-18 | 38.66 | 53.12 | 70.72 |
| Tile2Vec (Jean et al., 2019) | ResNet-18 | 40.60 | 52.22 | 65.46 |
| Contrastive (Chen et al., 2020a) | ResNet-18 | 54.52 | 64.94 | 80.56 |
| SimCLR (Chen et al., 2020a) | ResNet-18 | 66.70 | 75.94 | 81.24 |
| MoCo (He et al., 2020) | ResNet-18 | 57.64 | 65.70 | 81.02 |
| Ours | ResNet-18 | **67.62** | **76.52** | **81.78** |
| Random | ResNet-50 | 21.28 | 26.82 | 41.80 |
| ImageNet | ResNet-50 | 32.52 | 40.64 | 57.22 |
| Contrastive (Chen et al., 2020a) | ResNet-50 | 57.00 | 67.76 | 76.30 |
| SimCLR (Chen et al., 2020a) | ResNet-50 | 64.41 | 72.94 | 79.62 |
| MoCo (He et al., 2020) | ResNet-50 | 55.32 | 62.28 | 82.42 |
| Ours | ResNet-50 | **71.90** | **77.62** | **84.44** |

**Table 5**
Segmentation results on the DeepGlobe Dataset (Demir et al., 2018).

| Weights | ResNet-18 | | ResNet-50 | |
|---|---|---|---|---|
| | OA | mIoU | OA | mIoU |
| Random | 81.09 | 55.38 | 80.81 | 54.42 |
| ImageNet | 83.27 | 61.95 | 82.27 | 59.31 |
| Tile2Vec (Jean et al., 2019) | 80.50 | 56.93 | – | – |
| Contrastive (Chen et al., 2020a) | 85.25 | 64.85 | 86.06 | **68.46** |
| SimCLR (Chen et al., 2020a) | 85.65 | 66.15 | 83.80 | 63.97 |
| MoCo (He et al., 2020) | 84.79 | 65.28 | 85.07 | 66.17 |
| Ours | **86.11** | **67.07** | **86.58** | 67.87 |

**Table 6**
Results on the ADVANCE dataset (Hu et al., 2020a), values displayed in %.

| Model | Imagery | Audio | Precision | Recall | $F_1$ |
|---|---|---|---|---|---|
| Audio Baseline (Hu et al., 2020a) | ✗ | ✔ | 30.46 | 32.99 | 28.99 |
| Visual Baseline (Hu et al., 2020a) | ✔ | ✗ | 74.05 | 72.79 | 72.85 |
| AV Baseline (Hu et al., 2020a) | ✔ | ✔ | 75.25 | 74.79 | 74.58 |
| Ours (ResNet-18) | ✗ | ✔ | 37.91 | 38.36 | 37.69 |
| Ours (ResNet-18) | ✔ | ✗ | **87.09** | **87.07** | **86.92** |
| Ours (ResNet-18) | ✔ | ✔ | **89.59** | **89.52** | **89.50** |
| Ours (ResNet-50) | ✗ | ✔ | **39.13** | **39.96** | **39.01** |
| Ours (ResNet-50) | ✔ | ✗ | 83.97 | 83.88 | 83.84 |
| Ours (ResNet-50) | ✔ | ✔ | 88.90 | 88.85 | 88.83 |

**Table 7**
Results of the ablation study, values displayed in % except for the last row.

| Benchmark | Metric | Naive TL | | Contrastive loss | | Batch TL | |
|---|---|---|---|---|---|---|---|
| | | RN-18 | RN-50 | RN-18 | RN-50 | RN-18 | RN-50 |
| UC Merced (Yang and Newsam, 2010) | Accuracy | 85.14 | 77.43 | 86.48 | 88.19 | 90.19 | 89.71 |
| RESISC45 (Cheng et al., 2017) | Accuracy | 76.11 | 72.15 | 80.65 | 82.41 | 81.71 | 84.88 |
| AID (Xia et al., 2017) | Accuracy | 78.70 | 75.64 | 77.18 | 81.08 | 81.78 | 84.44 |
| DeepGlobe (Demir et al., 2018) | Accuracy | 83.96 | 85.40 | 80.72 | 85.96 | 86.11 | 86.58 |
| | mIoU | 63.14 | 65.18 | 57.26 | 67.28 | 67.07 | 67.87 |
| ADVANCE (Hu et al., 2020a) | F-Score | 88.51 | 87.61 | 79.42 | 80.84 | 89.46 | 88.83 |
| Cross-modal retrieval | Recall @ 100 | 18.59 | 13.41 | 29.12 | 28.35 | 19.01 | 15.28 |
| | Median rank | 749 | 951 | 565 | 580 | 744 | 836 |

The visual quality of the prediction results (see Fig. 5) varies a lot between the different evaluations. Small structures like scattered houses are not captured well by the methods that have never seen aerial imagery before (Random, ImageNet). The self-supervised methods trained on aerial imagery on the other hand have no issues picking up these structures.

### 5.4. Audiovisual scene classification

One application that has not received too much attention from the research community is audiovisual scene classification, where locally sourced audio data is combined with overhead imagery. Given that our framework exploits these very two modalities as well, we also include this task as a possible downstream task in our experiments. The *ADVANCE Dataset* (Hu et al., 2020a) poses a benchmark for audiovisual scene classification, and the accompanying research is a large source of inspiration for our work. On this dataset, our model outperforms the baseline set in Hu et al. (2020a) by a large margin, as can be seen in Table 6. These results suggest that for this task, self-supervised training on a large dataset beats direct, supervised training on a smaller dataset.

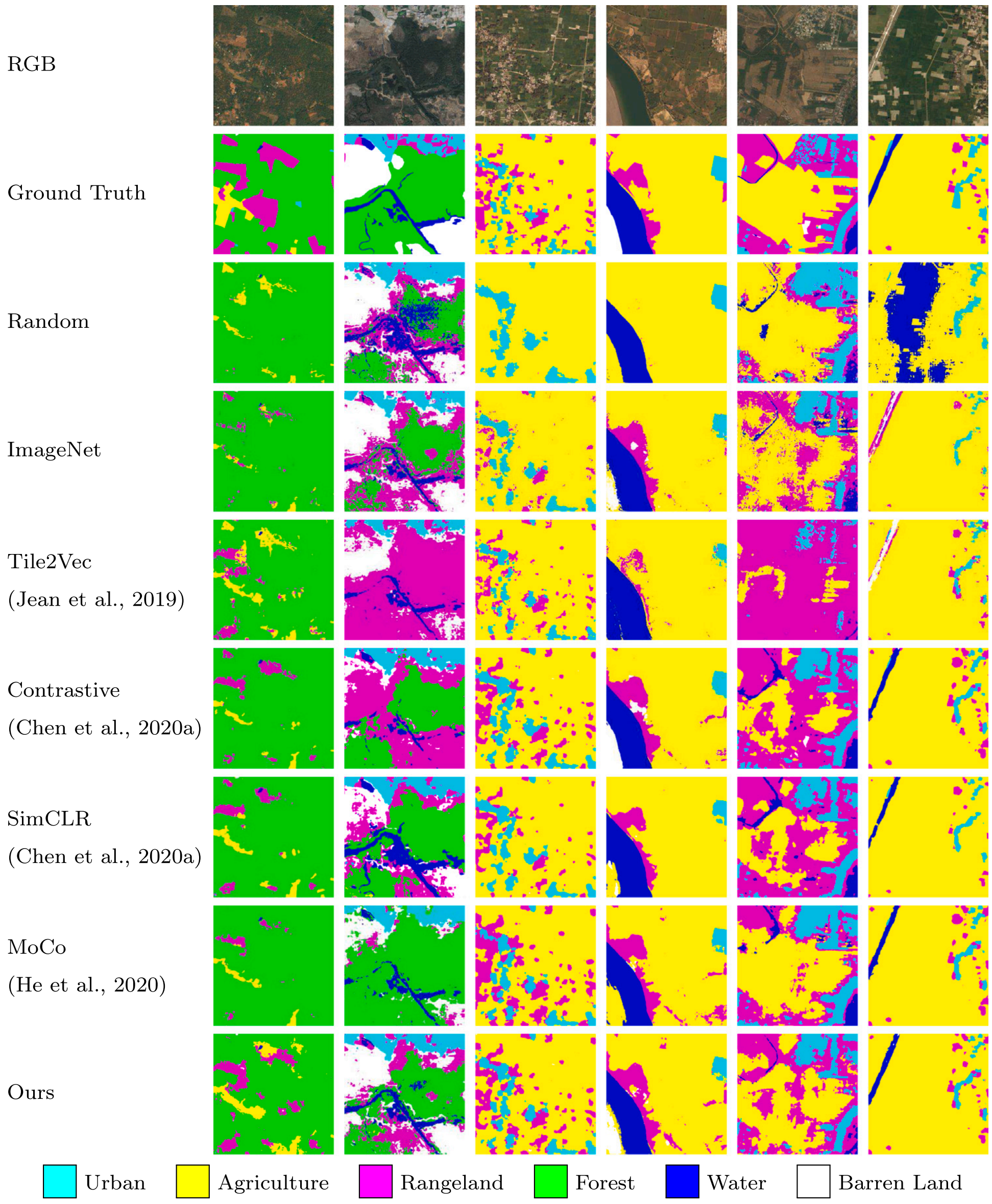

**Fig. 5.** Predictions of the different models on randomly selected validation tiles from the DeepGlobe Land Cover Classification dataset (Demir et al., 2018). The ResNet-50 version was used, except for Tile2Vec, where only ResNet-18 weights are available. Best viewed in color. (For interpretation of the references to color in this figure legend, the reader is referred to the web version of this article.)

### 5.5. *Cross-modal retrieval*

As a final application of our pretrained models, we evaluate the task of cross-modal retrieval. Given an input image, we try to predict the corresponding audio sample by retrieving the closest audio samples in the shared embedding space. Good performance in this task should imply high semantic similarity for neighboring points in this space.

It turns out that this task is really hard to perform on the given dataset. To understand this difficulty, imagine seeing an overhead image of city streets, which needs to be matched to exactly one out of hundreds of audio clips containing car and traffic sounds. This explains why in quantitative evaluations, the scores for our models look rather low. For the ResNet-18 model, 19.01% of all testing samples had the correct audio sample among the top 100 retrievals, while the median rank of the correct audio clip was at 744. The model based on ResNet-50 scores a bit lower on these metrics, reaching 15.28% and 836, respectively.

To put the retrieval results into perspective, we asked participants to assess the model performance in a kind of "Turing Test". In this human evaluation, we mixed up two kinds of sound-image pairs (35 pairs for each, 70 in total). The first kind is an image paired with the original sound while the other one is an image paired with the top-1 audio retrieved. These predicted pairs do not share the same overhead image. Given 70 pairs each, 15 participants were then asked to answer "Was the sound clip recorded somewhere within the image?". Then, we calculated the percentage of "Yes" answers for each kind of pair. For true pairings, the participants correctly answered "Yes" for 71.6% (±12.1%) of the samples. Surprisingly, the participants considered nearly the same proportion (69.5% ±13.1%) of the pairings suggested by our model to be true pairings. This validates the quality of retrieved sounds.

### 5.6. *Ablation study*

Finally, we conduct an ablation study to provide evidence that our Batch Triplet Loss function actually improves the quality of the learned representations over the other loss functions. Therefore, we compare the performance of models trained with our Batch Triplet Loss to models trained with plain Triplet Loss and the Contrastive Loss used in recent methods like SimCLR (Chen et al., 2020a).

Table 7 shows the results for the ablation study. First, and most importantly, we notice that Batch Triplet Loss outperforms the underlying naive Triplet Loss in all benchmarks. What is more, it also outperforms the strong competitor given by the Contrastive Loss in all tasks with the exception of the retrieval task, where Contrastive Loss outperforms the Triplet Loss-based models by a large margin. This suggests that batch triplet loss encourages the model to learn strongly discriminative features that are useful for classification and segmentation tasks, while contrastive learning is more suited for metric learning and clustering of the data.

## 6. Conclusion

With this work, we showed how the recent ideas in self-supervised learning can contribute to the improvement of deep learning models in remote sensing. By exploiting the strong connections between audio and imagery, our models can learn semantic representations of both modalities, without the need for a laborious manual annotation process. The resulting models outperform competing methods on a number of benchmark datasets, covering the tasks of aerial image classification, audiovisual scene classification, aerial image segmentation and cross-modal retrieval.

We hope that by making our code and pretrained weights available, further research on aerial imagery can profit directly from this pretraining method.

Further, the multimodal dataset that we built should open up interesting possibilities for further research in this direction, including more sophisticated multimodal representation learning methods.

## Declaration of competing interest

The authors declare that they have no known competing financial interests or personal relationships that could have appeared to influence the work reported in this paper.

## Data availability

The SoundingEarth dataset is available at https://zenodo.org/record/5600379. The accompanying code is available at https://github.com/khdlr/SoundingEarth.

## Acknowledgments

This research would not have been possible without the countless contributors to the radio aporee ::: maps project.

Further, we acknowledge Google for providing imagery from Google Earth for research purposes.

This work is supported by the German Federal Ministry of Education and Research (BMBF) in the framework of the international future AI lab “AI4EO – Artificial Intelligence for Earth Observation: Reasoning, Uncertainties, Ethics and Beyond” (grant number: 01DD20001), by the German Federal Ministry for Economic Affairs and Climate Action in the framework of the “national center of excellence ML4Earth” (grant number: 50EE2201C), by the Fundamental Research Funds for the Central Universities, China, by the Research Funds of Renmin University of China (NO. 2021030200), and by the Beijing Outstanding Young Scientist Program (NO. BJJWZYJH012019100020098), Public Computing Cloud, Renmin University of China.